\documentclass[11pt,a4paper]{article}

\usepackage[T1]{fontenc}
\usepackage{lmodern}            
\usepackage[scaled=0.85]{beramono} 
\usepackage{microtype}          

\usepackage[
  hmargin=3.8cm,
  vmargin=3cm
]{geometry}

\usepackage{parskip}            
\usepackage{enumitem}           
\setlist{
  topsep=0.5em,
  itemsep=0.2em,
  parsep=0pt
}

\usepackage{fancyhdr}
\fancypagestyle{plain}{
  \fancyhf{}
  
  \fancyfoot[C]{\small\thepage}
}

\usepackage{titlesec}

\titleformat{\section}
  {\large\sffamily\bfseries}
  {\thesection}{0.75em}{}

\titleformat{\subsection}
  {\normalsize\sffamily\bfseries}
  {\thesubsection}{0.75em}{}

\titlespacing*{\section}{0pt}{2.5ex}{1.2ex}
\titlespacing*{\subsection}{0pt}{2ex}{1ex}

\usepackage{xcolor}
\definecolor{accent}{HTML}{2563EB}    
\definecolor{textcodered}{rgb}{0.55,0.1,0.1}

\usepackage[
  colorlinks = true,
  urlcolor = accent,
  citecolor = accent,
  linkcolor = accent
]{hyperref}

\usepackage{graphicx}
\usepackage{booktabs}
\usepackage{caption}
\usepackage{tikz}
\usetikzlibrary{arrows.meta, positioning, shapes.geometric, fit, backgrounds}
\usepackage{float}  
\usepackage{wrapfig}

\usepackage[round,authoryear]{natbib}

\newcommand{\mono}[1]{\texttt{\color{textcodered}#1}}
\newcommand{\mjlab}{\mono{mjlab}}

\newcommand{\mujoco}{\textsf{MuJoCo}}

\usepackage{mdframed}
\renewenvironment{abstract}{%
  \begin{mdframed}[
    linewidth=0pt,
    backgroundcolor=black!4,
    innertopmargin=12pt,
    innerbottommargin=12pt,
    innerleftmargin=14pt,
    innerrightmargin=14pt,
    roundcorner=3pt,
    skipabove=\baselineskip,
    skipbelow=\baselineskip
  ]
  \noindent\small
}{%
  \end{mdframed}
}

\title{%
  \vspace{-1.5cm}%
  {\Huge\texttt{\textbf{mjlab}}}\\[6pt]
  {\large\textsf{A Lightweight Framework for GPU-Accelerated Robot Learning}}\\[3pt]
  {\normalsize\href{https://github.com/mujocolab/mjlab}{\textsf{github.com/mujocolab/mjlab}}}%
  \vspace{-0.5cm}%
}
\author{%
  \normalsize Kevin Zakka\textsuperscript{1} \quad
  Qiayuan Liao\textsuperscript{1} \quad
  Brent Yi\textsuperscript{1} \quad
  Louis Le Lay\textsuperscript{2}\\[2pt]
  Koushil Sreenath\textsuperscript{1} \quad
  Pieter Abbeel\textsuperscript{1}\\[4pt]
  \small\textsuperscript{1}UC Berkeley \quad
  \textsuperscript{2}Sorbonne University
}
\date{}

\begin{document}

\maketitle
\thispagestyle{plain}
\vspace{-2em}

\begin{center}
  \includegraphics[width=\textwidth]{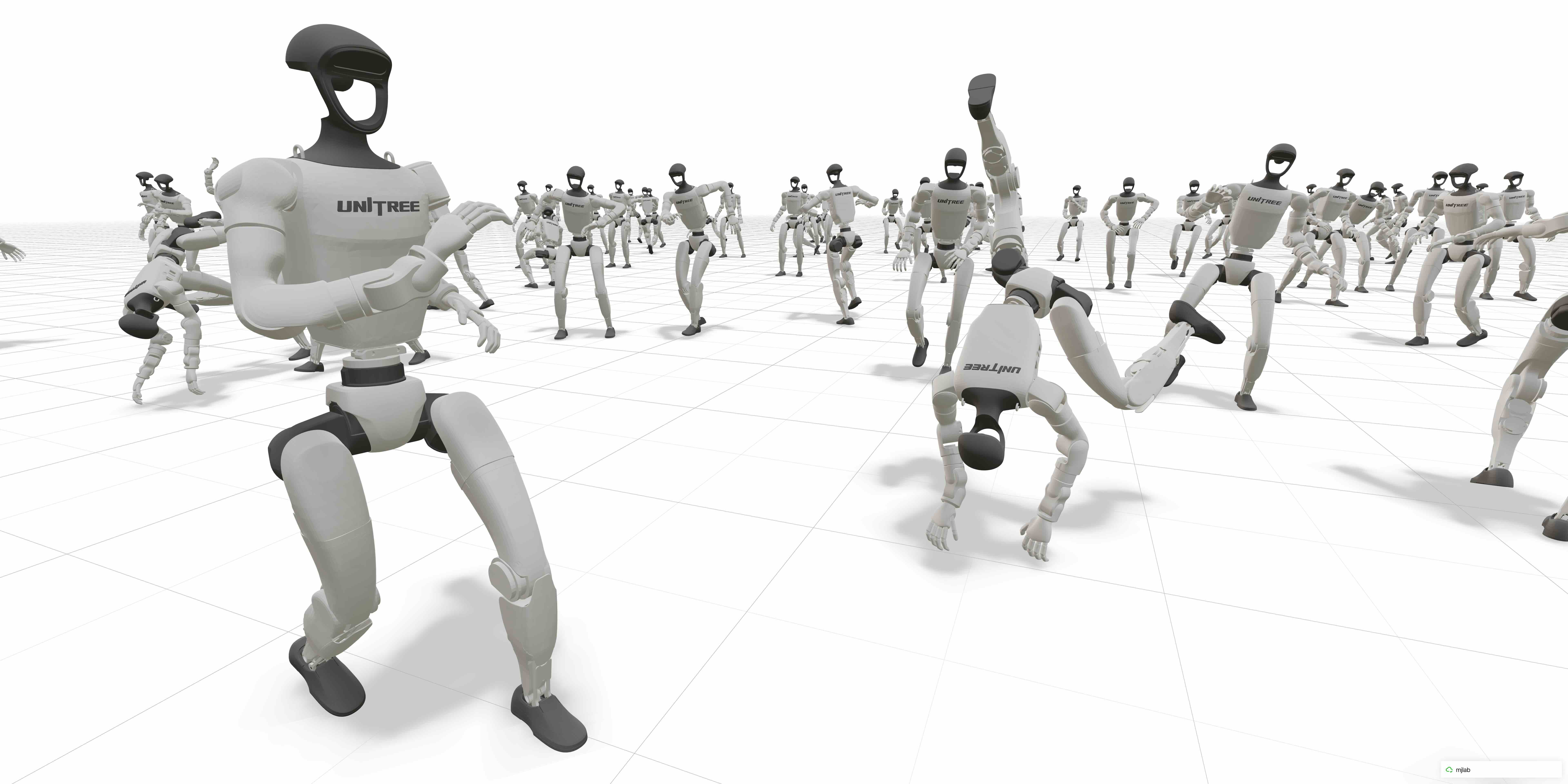}
  \captionof{figure}{Unitree G1 humanoids performing a dance trained with the motion tracking pipeline from BeyondMimic~\citep{liao2025beyondmimic}. Thousands of parallel environments are simulated on a single GPU and visualized in \mjlab{}'s web-based Viser viewer.}
  \label{fig:teaser}
\end{center}

\begin{abstract}%
We present \mjlab{}, a lightweight, open-source framework for robot learning that combines GPU-accelerated simulation with composable environments and minimal setup friction.
\mjlab{} adopts the manager-based API introduced by Isaac Lab~\citep{mittal2025isaaclab}, where users compose modular building blocks for observations, rewards, and events, and pairs it with \mujoco{} Warp~\citep{mujocowarp2025} for GPU-accelerated physics. The result is a framework installable with a single command, requiring minimal dependencies, and providing direct access to native \mujoco{} data structures. \mjlab{} ships with reference implementations of velocity tracking, motion imitation, and manipulation tasks.
\end{abstract}

\section{Introduction}

Reinforcement learning has become a powerful tool for training robot controllers in simulation and transferring them to real hardware. The fidelity of this sim-to-real pipeline hinges on getting simulation details right: actuator dynamics, contact modeling, sensor noise, and domain randomization all require careful implementation. Middleware that handles this infrastructure, ideally written and vetted by the people closest to the simulator, frees researchers to focus on what matters most: shaping reward functions, designing curriculums, and iterating on policies.
\newpage
Several frameworks address this need. Isaac Lab~\citep{mittal2025isaaclab} provides a comprehensive, GPU-accelerated platform with a rich manager-based API for composing RL environments. However, it requires the Omniverse runtime, which adds installation complexity and startup latency. Its physics engine, PhysX, was closed-source until recently, making low-level debugging and introspection difficult. \mujoco{} Playground~\citep{zakka2025mujocoplayground} takes the opposite approach: minimal abstractions and monolithic environment definitions that are easy to hack and quick to prototype. This design excels for one-off experiments, but code duplication across robots and tasks accumulates quickly, making multi-robot or multi-task codebases difficult to maintain. There remains a gap for a framework that is both lightweight and built on a proven orchestration API with access to best-in-class physics.

\mjlab{} fills this gap. It adopts Isaac Lab's manager-based design, where users compose self-contained building blocks for observations, rewards, events, and commands, and pairs it with \mujoco{} Warp for GPU-accelerated physics simulation. The result is a framework with minimal dependencies, fast startup, direct access to native \mujoco{} model and data structures, and a PyTorch~\citep{Ansel_PyTorch_2_Faster_2024}-native interface for policy training. \mjlab{} ships with three robot morphologies and three reference tasks. The remainder of this report describes the design philosophy~(Section~\ref{sec:philosophy}), architecture~(Section~\ref{sec:architecture}), core components~(Section~\ref{sec:components}), the manager-based API~(Section~\ref{sec:managers}), shipped tasks~(Section~\ref{sec:tasks}), and software design~(Section~\ref{sec:software}).



\section{Design Philosophy and Scope}
\label{sec:philosophy}

\mjlab{} is designed around three core engineering commitments: (1)~minimal installation friction, (2)~transparent and inspectable physics, and (3)~tight integration with the MuJoCo ecosystem. These commitments motivate deliberate trade-offs in implementation and scope, which we make explicit below.

\vspace{-0.5em}
\paragraph{Physics backend.}
\mjlab{} targets a single physics stack, MuJoCo Warp, to prioritize simulation transparency and debuggability. The framework exposes MuJoCo-native \mono{MjModel} and \mono{MjData} structures for direct inspection and state access. A non-goal is cross-simulator portability~\citep{Amazon_FAR_and_Abbeel_Holosoma}; \mjlab{} favors precise control and interpretability over backend generality.

\vspace{-0.5em}
\paragraph{Sensing and rendering.}
\mjlab{} provides ray-cast, depth, and RGB camera sensors for geometric and visual perception. Photorealistic rendering is out of scope. This does not preclude vision-based policies: a common approach is to train privileged policies using full state, then distill into vision-based controllers using external rendering.

\vspace{-0.5em}
\paragraph{Environment composition.}
\mjlab{} adopts Isaac Lab's manager-based API, where environments are built from reusable, composable components rather than monolithic definitions. Modular terms for observations, rewards, events, and curricula reduce code duplication across tasks. The implementation remains fully MuJoCo-native; users work directly with MuJoCo models and conventions.

\vspace{-0.5em}
\paragraph{Scope and extensibility.}
\mjlab{} provides infrastructure for rigid-body robot learning that is intended to be extended to custom robots, tasks, sensors, and actuators. The framework includes three example robots and tasks that demonstrate use of the API.

Together, these choices reflect a focus on fast iteration, transparent physics, and MuJoCo-native workflows, rather than feature completeness or broad simulator support.

\section{Architecture}
\label{sec:architecture}

\begin{wrapfigure}{r}{0.52\textwidth}
\centering
\vspace{-1em}
\scalebox{0.72}{
\begin{tikzpicture}[
  node distance=0.8cm and 1cm,
  box/.style={rectangle, draw, rounded corners=3pt, minimum width=2.8cm, minimum height=1cm, align=center, font=\small\sffamily, inner sep=5pt},
  entity/.style={box, fill=blue!8, minimum width=2.6cm, minimum height=1.5cm},
  process/.style={box, fill=gray!12},
  gpu/.style={box, fill=orange!15, minimum width=3.2cm},
  manager/.style={rectangle, draw, rounded corners=2pt, minimum width=1.3cm, minimum height=0.6cm, align=center, font=\scriptsize\sffamily, fill=green!10},
  arrow/.style={-{Stealth[length=2mm]}, thick}
]

\node[entity] (e1) {Entity 1\\[2pt]{\scriptsize (G1 robot)}\\[2pt]{\scriptsize\color{gray} MJCF + configs}};
\node[entity, right=0.6cm of e1] (e2) {Entity 2\\[2pt]{\scriptsize (Cube)}\\[2pt]{\scriptsize\color{gray} MJCF}};
\node[entity, right=0.6cm of e2] (e3) {Entity 3\\[2pt]{\scriptsize (Terrain)}\\[2pt]{\scriptsize\color{gray} Python}};

\node[process, below=1.3cm of e2] (mjspec) {MjSpec};

\draw[arrow] (e1.south) -- ++(0,-0.4) -| ([xshift=-0.2cm]mjspec.north);
\draw[arrow] (e2.south) -- (mjspec.north);
\draw[arrow] (e3.south) -- ++(0,-0.4) -| ([xshift=0.2cm]mjspec.north);

\node[process, below=0.7cm of mjspec] (mjmodel) {MjModel};
\draw[arrow] (mjspec) -- (mjmodel);

\node[gpu, below=0.7cm of mjmodel] (warp) {MuJoCo Warp\\[-2pt]{\scriptsize (GPU, $N$ worlds)}};
\draw[arrow] (mjmodel) -- (warp);

\node[process, below=0.7cm of warp, minimum width=6.5cm, minimum height=2.5cm] (env) {};
\node[font=\small\sffamily\bfseries, anchor=north] at ([yshift=-0.1cm]env.north) {ManagerBasedRlEnv};

\node[manager, below=0.7cm of env.north west, xshift=1cm] (obs) {Obs};
\node[manager, right=0.2cm of obs] (act) {Action};
\node[manager, right=0.2cm of act] (rew) {Reward};
\node[manager, right=0.2cm of rew] (term) {Term};
\node[manager, below=0.2cm of obs] (event) {Event};
\node[manager, right=0.2cm of event] (cmd) {Cmd};
\node[manager, right=0.2cm of cmd] (curr) {Curric};
\node[manager, right=0.2cm of curr] (metrics) {Metrics};

\draw[arrow] (warp) -- (env);

\node[process, below=0.6cm of env] (rslrl) {RSL-RL};
\draw[arrow] (env) -- (rslrl);

\node[box, below=0.6cm of rslrl, fill=blue!15] (policy) {Trained Policy};
\draw[arrow] (rslrl) -- (policy);

\end{tikzpicture}
}
\vspace{-0.5em}
\caption{Entities are composed into an MjSpec, compiled, and transferred to MuJoCo Warp for GPU simulation. The ManagerBasedRlEnv orchestrates the MDP; RSL-RL handles training.}
\label{fig:architecture}
\vspace{-1em}
\end{wrapfigure}
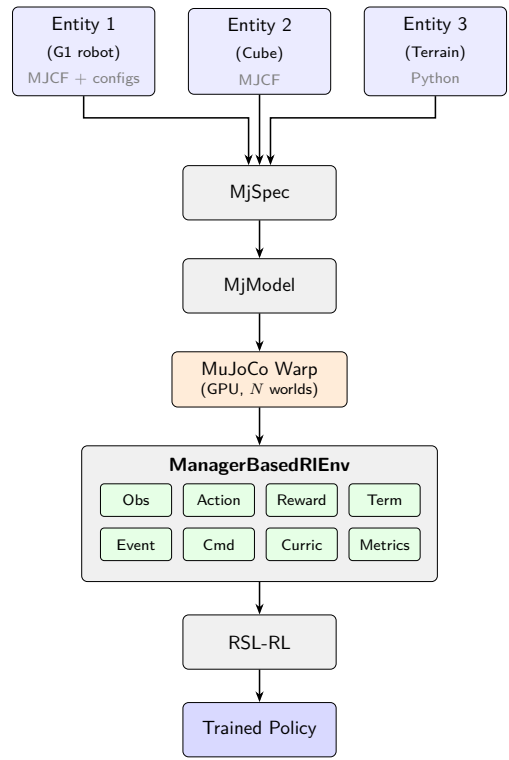

\textbf{\mujoco{} Warp.}
\mujoco{} Warp \citep{mujocowarp2025} (\href{https://mujoco.readthedocs.io/en/stable/mjwarp/index.html}{docs}) is a GPU-accelerated backend for \mujoco{} built on \href{https://nvidia.github.io/warp/}{NVIDIA Warp}~\citep{warp2022}. It preserves \mujoco{}'s~\citep{todorov2012mujoco} familiar \mono{MjModel}/\mono{MjData} paradigm: \mono{MjModel} holds the static kinematic and dynamic description of the scene, while \mono{MjData} carries the time-varying simulation state. The key addition is a leading \emph{world} dimension: a single \mono{MjData} object holds the state of $N$ independent simulation instances in parallel, enabling thousands of environments to be stepped in parallel. Model parameters are shared across all worlds by default, and individual fields can be expanded to vary per-world when domain randomization requires it. \mjlab{} further captures the simulation step as a CUDA graph: the kernel execution sequence is recorded once and replayed on subsequent calls, eliminating CPU-side dispatch overhead.

\textbf{Scene pipeline.}
\mjlab{}~constructs scenes by composing entity descriptions defined via MJCF into a single \href{https://mujoco.readthedocs.io/en/stable/programming/modeledit.html}{\mono{MjSpec}}. This specification is compiled into an \mono{MjModel} on the CPU, then transferred to the GPU via \mujoco{} Warp's \mono{put\_model} and \mono{put\_data} routines.

\textbf{Manager-based paradigm.}
On top of the simulation layer, \mjlab{} adopts the manager-based environment design introduced by Isaac Lab. Users define their environment by composing small, self-contained \emph{terms}---reward functions, observation computations, domain randomization events---and register them with the appropriate manager. Each manager handles the lifecycle of its terms: calling them at the right point in the simulation loop, aggregating their outputs, and exposing diagnostics. Terms can be plain functions for stateless computations or classes that inherit from \mono{ManagerTermBase} when they need to cache expensive setup (e.g., resolving regex patterns to joint indices at initialization) or maintain state across steps with a per-episode \mono{reset()} hook. This abstraction introduces overhead compared to writing monolithic step functions, but benefits in modularity, testability, and rapid iteration outweigh the cost, a lesson reinforced by our experience with the more minimal \mujoco{} Playground API, where the absence of such structure made environments harder to maintain and extend.

\section{Components}
\label{sec:components}

\mjlab{} provides a set of core components that managers operate on during the simulation lifecycle.

\textbf{Entities.}
An entity represents a physical object in the scene, such as robots, manipulated objects, or static fixtures. While Isaac Lab splits this concept across multiple specialized classes (\mono{Articulation}, \mono{RigidObject}, \mono{DeformableObject}, \mono{RigidObjectCollection}, and others), \mjlab{} uses a single \mono{Entity} class that covers all variations through two independent properties: base type (fixed or floating) and articulation (with or without joints). Runtime queries such as \mono{is\_fixed\_base} and \mono{is\_articulated} let downstream code branch where needed without requiring a class hierarchy. Each entity maintains an \mono{EntityData} structure that aggregates kinematic quantities such as body positions, orientations, and velocities, making them available to observation and reward terms without redundant computation.

\textbf{Sensors.}
Sensors in \mjlab{} follow a layered design. Users can wrap sensors already defined in their MJCF files and read them through \mujoco{}'s native \mono{sensordata} buffer, or define them entirely in Python configuration. On top of this native layer, \mjlab{} provides custom sensor types that go beyond what \mujoco{} offers: a BVH-accelerated ray-cast sensor that supports grid and pinhole ray patterns and returns per-ray distances, hit positions, and surface normals; a contact sensor that tracks collision forces between body pairs with air-time and force-history bookkeeping; and a GPU-accelerated camera sensor that renders RGB and depth across all worlds in parallel and supports both perspective and orthographic projection. All sensors implement a common interface with per-step caching and optional debug visualization.

\textbf{Actuators.}
Actuators follow the same layered philosophy. Users can wrap actuators defined in MJCF, preserving the exact behavior specified in the model file, or create \mujoco{}-native actuator types (motor, position, velocity, muscle) from configuration. Beyond these native options, \mjlab{} provides custom implementations that compute torques on the GPU outside of \mujoco{}: an ideal PD controller, a DC motor model with velocity-dependent torque saturation, and a learned MLP actuator for capturing hardware-specific dynamics from data. All three tiers coexist within the same entity. Actuation delay, common in real robots, is modeled by a wrapper class that buffers control signals and replays them with a latency quantized to the physics timestep.

\begin{figure}[t]
  \centering
  \includegraphics[width=\textwidth]{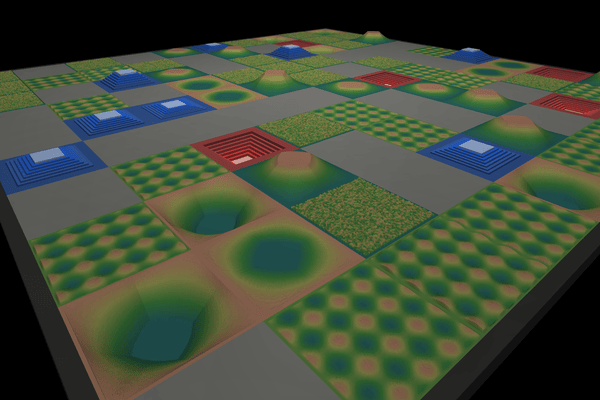}
  \caption{A composite terrain grid generated by \mjlab{}. Sub-terrain patches include flat ground, stairs, sloped pyramids, heightfield noise, and wave patterns. Difficulty increases along one axis for curriculum-based training.}
  \label{fig:terrain}
\end{figure}

\textbf{Terrain.}
The terrain module generates ground geometry for locomotion training. Users can spawn a flat ground plane or compose a grid of sub-terrain patches, each drawn from a library of terrain types. Two families are provided: primitive terrains built from \mujoco{} box geoms (flat surfaces, stairs, stepping stones, narrow beams, and random-height grids) and heightfield terrains for smoother, continuous profiles (sloped pyramids, Perlin noise, and sinusoidal waves). Every terrain type accepts a difficulty parameter that scales its features---step height, slope angle, noise amplitude---from trivial to challenging. The grid generator supports two modes: random, where each patch is sampled independently, and curriculum, where rows encode increasing difficulty so that the curriculum manager can advance robots to harder terrain as performance improves. Users can subclass the base terrain configuration to define custom terrain types (Figure~\ref{fig:terrain}).

\textbf{Viewer.}
\mjlab{} supports two visualization backends. The native \mujoco{} viewer provides the full \mono{simulate} experience---pause and resume, contact point and force visualization, interactive perturbations, and more. For headless servers or remote machines, \mjlab{} offers a web-based viewer built on Viser~\citep{yi2025viser} that supports pause, resume, and contact visualization in the browser.

\section{Manager-Based API}
\label{sec:managers}

\mjlab{} environments implement the Gym~\citep{brockman2016gym} interface, a standard API for defining Markov decision processes (MDPs). Environments expose two methods: \mono{reset()} initializes a new episode and returns the first observation, while \mono{step()} applies an action and returns the next observation, reward, and termination signal. Each call to \mono{step()} proceeds through a fixed pipeline:
\begin{enumerate}[leftmargin=*, itemsep=2pt, parsep=0pt, topsep=4pt]
  \item \textbf{Act.} The action manager updates the action history and splits the policy output into per-term slices.
  \item \textbf{Simulate.} For each of $d$ decimation sub-steps: apply actuator commands, write controls to the simulation, advance physics, and update entity state.
  \item \textbf{Terminate.} The termination manager evaluates stop conditions and flags environments for reset.
  \item \textbf{Reward.} The reward manager computes and aggregates weighted reward terms.
  \item \textbf{Reset.} Flagged environments are reset: the curriculum manager updates difficulty, entity states are reinitialized, and the event manager applies reset hooks.
  \item \textbf{Command.} The command manager generates or resamples goal signals.
  \item \textbf{Apply Events.} The event manager applies periodic events (e.g., external pushes).
  \item \textbf{Observe.} The observation manager computes observations for the next policy query.
\end{enumerate}

This manager-based orchestration follows the design introduced by Isaac Lab~\citep{mittal2025isaaclab}, where each manager encapsulates a specific aspect of the MDP and users compose environments by registering modular terms.

\textbf{Action manager.}
The action manager receives the policy's output tensor, splits it according to the registered action terms, and routes each slice to the appropriate entity's actuators. It supports action clipping and tracks the two most recent actions, which observation and reward terms can reference (e.g., to compute an action rate penalty). Shipped action terms include joint-space position and velocity targets as well as a differential inverse kinematics term that converts task-space pose commands into joint displacements via damped least-squares, with optional null-space posture regularization and soft joint-limit avoidance.

\textbf{Termination manager.}
The termination manager evaluates a set of Boolean conditions each step---contact with illegal bodies, joint limit violations, timeout---and flags environments for reset. It also monitors simulation state for NaN and Inf values, catching numerical instabilities before they propagate. When an instability is detected, \mjlab{} captures a rolling buffer of recent states that users can replay in an interactive viewer to diagnose the failure. The observation and reward managers additionally report which specific terms produced invalid values, narrowing the search from ``something exploded'' to the exact offending computation.

\textbf{Reward manager.}
Reward terms are functions or stateful classes that return per-environment scalars. The manager computes a weighted sum of all active terms, automatically scaling each by the environment's control timestep to make reward magnitudes invariant to the simulation frequency. Per-term episodic sums are tracked and logged, making it straightforward to diagnose which terms dominate training. Terms that produce NaN or Inf values are detected and reported individually.

\textbf{Curriculum manager.}
The curriculum manager adjusts training conditions based on policy performance. Examples include scaling reward term weights, widening command velocity ranges, or advancing robots to harder terrain rows as success rates improve.

\textbf{Event manager.}
Events are hooks that execute at specific points in the environment lifecycle: at startup, on reset, or at fixed intervals during training. The most common use case is domain randomization. \mjlab{} provides a domain randomization module covering body, joint, geometry, camera, light, material, and tendon parameters, each with configurable operations (absolute, additive, multiplicative) and sampling distributions (uniform, log-uniform, Gaussian). When an event modifies a field of the \mujoco{} model, \mjlab{} expands that field from a shared value to a per-world array and rebuilds the CUDA capture graph to reflect the new memory layout. For inertial parameters, the module implements physically consistent randomization based on the pseudo-inertia parameterization of \citet{rucker2022smooth}, guaranteeing that randomized mass, center of mass, and rotational inertia remain fully physically consistent for any perturbation magnitude. The breadth of randomizable parameters, from link geometry and inertial properties to joint dynamics and actuator gains, combined with physical consistency, makes the module applicable beyond sim-to-real transfer to problems such as system identification and hardware co-design.

\textbf{Command manager.}
The command manager generates goal signals for the policy---velocity targets, tracking waypoints, or object poses---and resamples them at configurable intervals. Commands are exposed to the observation manager so the policy can condition on the current goal.

\textbf{Observation manager.}
Observation terms are composable functions that read from entity data, sensors, or commands and return tensors. The manager chains these terms through a configurable pipeline: raw values are optionally clipped, scaled, corrupted with noise, delayed by a configurable number of control steps, and concatenated with a history buffer. Multiple observation groups (e.g., \emph{policy} and \emph{critic}) can coexist, each with its own processing pipeline, enabling asymmetric actor-critic architectures.

\textbf{Metrics manager.}
The metrics manager computes and logs custom per-episode metrics that are not tied to any specific manager, such as task success rates or tracking errors, providing a unified interface for monitoring training progress.

\section{Tasks}
\label{sec:tasks}

\begin{figure}[t]
  \centering
  \includegraphics[width=\textwidth]{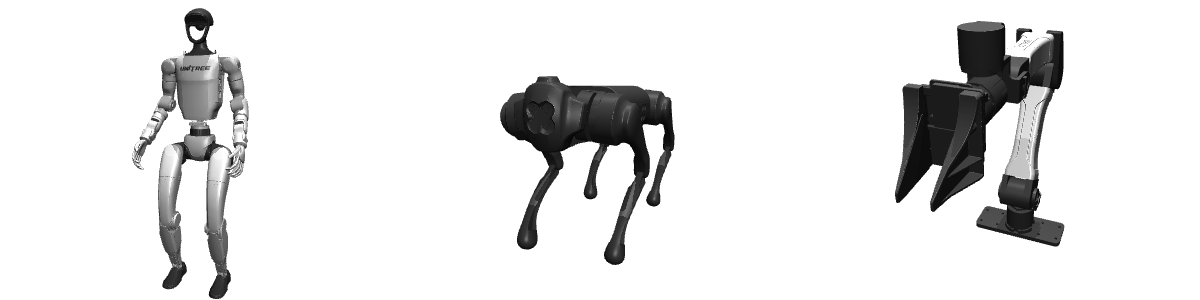}
  \caption{The three robot morphologies shipped with \mjlab{}: Unitree G1 humanoid (left), Unitree Go1 quadruped (center), and YAM robot arm (right). All models are adapted from \mujoco{} Menagerie~\citep{zakka2022menagerie}.}
  \label{fig:robots}
\end{figure}

\mjlab{} ships with three reference tasks spanning locomotion, whole-body control, and manipulation. The codebase is intentionally lean: three robot morphologies---a humanoid (\href{https://unitree.com/}{Unitree G1}), a quadruped (\href{https://unitree.com/}{Unitree Go1}), and a robot arm (\href{https://i2rt.com/collections/yam-arm}{YAM})---cover a broad range of research use cases while keeping the repository easy to navigate and maintain.

\subsection{Locomotion: Velocity Tracking}

The velocity tracking task trains a locomotion controller to follow linear and angular velocity commands on flat or rough terrain. The policy observes IMU readings, projected gravity, joint positions and velocities, the previous action, and the commanded twist. Reward terms encourage accurate velocity tracking while penalizing excessive body angular velocity, angular momentum, joint limit violations, action rate, and foot slip. Terrain variants range from flat ground to composite grids of stairs and rough heightfields, with terrain difficulty tied to the curriculum manager. Videos show a natural running gait on the G1 humanoid emerging from this reward structure in both simulation~[\href{https://x.com/kevin_zakka/status/1983965656220627176}{link}] and on real hardware~[\href{https://x.com/kevin_zakka/status/1984729094312116629}{link}].

\subsection{Whole-Body Control: Motion Imitation}

The motion imitation task trains a humanoid (Unitree G1) to track reference motion clips, implementing the DeepMimic~\citep{peng2018deepmimic} framework with extensions from BeyondMimic~\citep{liao2025beyondmimic}. The policy observes an anchor pose from the reference trajectory, base velocities, joint states, and the current action. Reward terms penalize deviations in global root position and orientation, relative body poses, and body velocities. A self-collision cost discourages interpenetration. A demonstration of a trained triple spinkick is available online~[\href{https://x.com/kevin_zakka/status/1976460408077812085}{link}].

\subsection{Manipulation: Cube Lifting}

The manipulation task trains a YAM robot arm to lift a cube to a target pose. The policy observes joint positions and velocities, the vector from end-effector to cube, the cube-to-goal error, and the previous action. A staged reward structure first guides the end-effector toward the cube, then rewards lifting the cube to the target height. Contact sensors on the end-effector and ground plane provide auxiliary signals. A video demonstration is available online~[\href{https://x.com/kevin_zakka/status/1995609144645353762}{link}].

\section{Software Design}
\label{sec:software}

A framework's long-term usability depends as much on how its code is organized as on what it can simulate. We made several deliberate departures from Isaac Lab's software patterns to reduce boilerplate, catch errors earlier, and lower the barrier to writing new tasks.

\textbf{PyTorch-native interface.}
\mujoco{} Warp stores simulation state in Warp arrays, while policy training typically happens in PyTorch. \mjlab{} bridges this gap with a \mono{TorchArray} abstraction: a zero-copy wrapper that exposes Warp arrays as PyTorch tensors. Users write rewards, observations, and custom logic in pure PyTorch without touching Warp kernels, and the framework handles memory sharing transparently.

\textbf{Instance-based configuration.}
Isaac Lab organizes environment parameters through deeply nested dataclass hierarchies: each new task inherits from a base config class and overrides fields via \mono{\_\_post\_init\_\_} mutations or field redeclarations. This pattern is fragile: \mono{\_\_post\_init\_\_} overrides can silently ignore constructor arguments, and a misspelled field name creates a new field rather than overriding the parent's, with no error at any stage. \mjlab{} replaces class-based configs with plain config \emph{instances}. Observation, reward, event, and other term configurations are held in typed dictionaries (\mono{dict[str, RewardTermCfg]}, etc.) rather than in bespoke dataclass subclasses. Creating a task variant is a matter of copying an existing config and mutating it, not defining a new class. This eliminates the inheritance pitfalls while making configs easier to read, diff, and compose programmatically.

\textbf{CLI-first configuration.}
Because configs are ordinary dataclass instances with typed fields, \mjlab{} can expose every parameter at the command line automatically via \mono{tyro}. Users override settings like reward weights, observation noise, simulation timestep with a single flag, without writing configuration files or subclassing.

\textbf{Co-located definitions.}
Isaac Lab separates configuration dataclasses from their implementations: all manager term configs live in a single \mono{manager\_term\_cfg.py}, while the corresponding classes are spread across other files. Configs must reference their implementation via a \mono{class\_type} field, adding indirection. \mjlab{} keeps each config dataclass in the same file as its implementation---\mono{ActionTermCfg} alongside \mono{ActionTerm}, for example---so related code is always collocated and navigation is straightforward.

\textbf{Static typing and testing.}
The entire codebase is statically type-checked with both \mono{pyright} and \mono{ty}, and ships a \mono{py.typed} marker for downstream consumers. A test suite of over 40 files covers managers, sensors, actuators, domain randomization, task execution, and more. 

\textbf{AI-assisted development.}
These design choices---typed configs, comprehensive tests, and a lean codebase---make \mjlab{} particularly amenable to AI coding agents. Tools such as Claude Code can type-check their own edits, run the test suite to verify correctness, and iterate autonomously with minimal human intervention. In practice, several recent contributions were implemented end-to-end by AI agents (e.g., \href{https://github.com/mujocolab/mjlab/pull/532}{PR~\#532}), suggesting that a well-structured codebase lowers the barrier for not only human contributions but also for automated ones.

\textbf{Minimal dependencies.}
A guiding constraint throughout \mjlab{}'s development was keeping installation friction-free. The framework avoids heavyweight dependencies, and is designed around \mono{uv}~\citep{uv2025}, a fast Python package manager that eliminates explicit environment setup. Users can go from a fresh clone to a running training job in seconds:
\begin{verbatim}
git clone https://github.com/mujocolab/mjlab && cd mjlab
uv run train Mjlab-Velocity-Flat-Unitree-G1 \
  --env.scene.num-envs 4096 \
  --agent.max-iterations 10_000
\end{verbatim}
\noindent Environment creation, installation, and dependency resolution happen automatically. Training uses RSL-RL~\citep{schwarke2025rslrl} for on-policy algorithms, with multi-GPU scaling available via \mono{torchrunx}~\citep{torchrunx2025}.

\section{Adoption}

\mjlab{} has been deployed in UC Berkeley's graduate robotics course (ME 292b/193b), where students new to reinforcement learning trained their first legged locomotion policies. The framework has been adopted by various open-source projects~[\href{https://github.com/MyoHub/mjlab_myosuite}{1}, \href{https://github.com/MarcDcls/mjlab_upkie}{2}] and featured in a tutorial from a popular YouTuber with over 1.5M subscribers~[\href{https://www.youtube.com/watch?v=FGnAeUXRZ4E}{link}].

\section{Acknowledgments}
We would like to thank Yuval Tassa, Erik Frey, Baruch Tabanpour, and Taylor Howell for their help with \mujoco{} and \mujoco{} Warp. We also thank Adam Allevato, Gr\'{e}goire Passault, Anoop Gadhrri, Sergi de la Muelas Acosta, Mustafa Haiderbhai, Haoyang Wang, Alphonsus Adu-Bredu, Bolaji Adesina, Tatsuki Tsujimoto, Vittorio Caggiano, Zahi Kakish, and Andrew Carr for their code contributions, and Robin Deits for finding and reporting bugs.

{\small
\bibliographystyle{plainnat}
\bibliography{main}

@article{mittal2025isaaclab,
  title={Isaac Lab: A GPU-Accelerated Simulation Framework for Multi-Modal Robot Learning},
  author={Mittal, Mayank and Roth, Pascal and Tigue, James and Richard, Antoine and Zhang, Octi and Du, Peter and others},
  journal={arXiv preprint arXiv:2511.04831},
  year={2025},
  url={https://arxiv.org/abs/2511.04831}
}

@article{schwarke2025rslrl,
  title={RSL-RL: A Learning Library for Robotics Research},
  author={Schwarke, Clemens and Mittal, Mayank and Rudin, Nikita and Hoeller, David and Hutter, Marco},
  journal={arXiv preprint arXiv:2509.10771},
  year={2025}
}

@article{yi2025viser,
    title={Viser: Imperative, web-based 3d visualization in python},
    author={Yi, Brent and Kim, Chung Min and Kerr, Justin and Wu, Gina and Feng, Rebecca and Zhang, Anthony and Kulhanek, Jonas and Choi, Hongsuk and Ma, Yi and Tancik, Matthew and Kanazawa, Angjoo},
    journal={arXiv preprint arXiv:2507.22885},
    year={2025}
}

@software{mujocowarp2025,
  title={{MuJoCo Warp}},
  author={{Google DeepMind} and {NVIDIA}},
  year={2025},
  url={https://github.com/google-deepmind/mujoco_warp}
}

@misc{warp2022,
  title        = {Warp: A High-performance Python Framework for GPU Simulation and Graphics},
  author       = {Miles Macklin},
  month        = {March},
  year         = {2022},
  note         = {NVIDIA GPU Technology Conference (GTC)},
  howpublished = {\url{https://github.com/nvidia/warp}}
}

@inproceedings{Ansel_PyTorch_2_Faster_2024,
author = {Ansel, Jason and Yang, Edward and He, Horace and Gimelshein, Natalia and Jain, Animesh and Voznesensky, Michael and Bao, Bin and Bell, Peter and Berard, David and Burovski, Evgeni and Chauhan, Geeta and Chourdia, Anjali and Constable, Will and Desmaison, Alban and DeVito, Zachary and Ellison, Elias and Feng, Will and Gong, Jiong and Gschwind, Michael and Hirsh, Brian and Huang, Sherlock and Kalambarkar, Kshiteej and Kirsch, Laurent and Lazos, Michael and Lezcano, Mario and Liang, Yanbo and Liang, Jason and Lu, Yinghai and Luk, CK and Maher, Bert and Pan, Yunjie and Puhrsch, Christian and Reso, Matthias and Saroufim, Mark and Siraichi, Marcos Yukio and Suk, Helen and Suo, Michael and Tillet, Phil and Wang, Eikan and Wang, Xiaodong and Wen, William and Zhang, Shunting and Zhao, Xu and Zhou, Keren and Zou, Richard and Mathews, Ajit and Chanan, Gregory and Wu, Peng and Chintala, Soumith},
booktitle = {29th ACM International Conference on Architectural Support for Programming Languages and Operating Systems, Volume 2 (ASPLOS '24)},
doi = {10.1145/3620665.3640366},
month = apr,
publisher = {ACM},
title = {{PyTorch 2: Faster Machine Learning Through Dynamic Python Bytecode Transformation and Graph Compilation}},
url = {https://docs.pytorch.org/assets/pytorch2-2.pdf},
year = {2024}
}

@inproceedings{todorov2012mujoco,
  title={Mujoco: A physics engine for model-based control},
  author={Todorov, Emanuel and Erez, Tom and Tassa, Yuval},
  booktitle={2012 IEEE/RSJ international conference on intelligent robots and systems},
  pages={5026--5033},
  year={2012},
  organization={IEEE}
}

@misc{liao2025beyondmimic,
  title={BeyondMimic: From Motion Tracking to Versatile Humanoid Control via Guided Diffusion},
  author={Qiayuan Liao and Takara E. Truong and Xiaoyu Huang and Yuman Gao and Guy Tevet and Koushil Sreenath and C. Karen Liu},
  year={2025},
  eprint={2508.08241},
  archivePrefix={arXiv},
  primaryClass={cs.RO},
  url={https://arxiv.org/abs/2508.08241},
}

@article{peng2018deepmimic,
  title={DeepMimic: Example-Guided Deep Reinforcement Learning of Physics-Based Character Skills},
  author={Peng, Xue Bin and Abbeel, Pieter and Levine, Sergey and van de Panne, Michiel},
  journal={ACM Transactions on Graphics},
  volume={37},
  number={4},
  year={2018}
}

@software{zakka2022menagerie,
  author = {Zakka, Kevin and Tassa, Yuval and {MuJoCo Menagerie Contributors}},
  month = sep,
  title = {{MuJoCo Menagerie: A collection of high-quality simulation models for MuJoCo}},
  year = {2022},
  url = {https://github.com/google-deepmind/mujoco_menagerie}
}

@misc{brockman2016gym,
  title={OpenAI Gym},
  author={Greg Brockman and Vicki Cheung and Ludwig Pettersson and Jonas Schneider and John Schulman and Jie Tang and Wojciech Zaremba},
  year={2016},
  eprint={1606.01540},
  archivePrefix={arXiv},
  primaryClass={cs.LG}
}

@software{uv2025,
  title={uv: An Extremely Fast Python Package and Project Manager},
  author={{Astral}},
  year={2025},
  url={https://docs.astral.sh/uv/}
}

@misc{zakka2025mujocoplayground,
  title={MuJoCo Playground},
  author={Kevin Zakka and Baruch Tabanpour and Qiayuan Liao and Mustafa Haiderbhai and Samuel Holt and Jing Yuan Luo and Arthur Allshire and Erik Frey and Koushil Sreenath and Lueder A. Kahrs and Carmelo Sferrazza and Yuval Tassa and Pieter Abbeel},
  year={2025},
  eprint={2502.08844},
  archivePrefix={arXiv},
  primaryClass={cs.RO},
  url={https://arxiv.org/abs/2502.08844},
}

@software{torchrunx2025,
  title = {torchrunx: Functional Distributed PyTorch Utility},
  author = {{Khandelwal, Apoorv and Curtin, Peter}},
  year = {2025},
  url = {https://torchrun.xyz/},
}

@article{rucker2022smooth,
  author={Rucker, Caleb and Wensing, Patrick M.},
  journal={IEEE Robotics and Automation Letters},
  title={Smooth Parameterization of Rigid-Body Inertia},
  year={2022},
  volume={7},
  number={2},
  pages={2771-2778},
  doi={10.1109/LRA.2022.3144517}
}

@software{Amazon_FAR_and_Abbeel_Holosoma,
author = {{Amazon FAR} and Abbeel, Pieter and Chen, Juyue and Duan, Rocky and Escontrela, Alejandro and Gandhi, Manan and Gundry, Samuel and Huang, Xiaoyu and Kanazawa, Angjoo and Lewicki, Tomasz and Li, Jiaman and Liu, Karen and Rosenthal, Clay and Seo, Younggyo and Sferrazza, Carlo and Shi, Guanya and Shih, Linda and Tseng, Jonathan and Wu, Zhen and Yang, Lujie and Yi, Brent and Zhang, Yuanhang},
title = {{Holosoma}}
}
}

\end{document}